\documentclass[10pt,twocolumn]{article}
\usepackage[T1]{fontenc}
\usepackage[margin=0.75in]{geometry}
\usepackage{booktabs}
\usepackage{graphicx}
\graphicspath{{./}}
\usepackage{amsmath}
\usepackage{amssymb}
\usepackage{url}
\usepackage[hidelinks]{hyperref}
\usepackage{microtype}
\usepackage{caption}
\title{Fusion Under Component Failure:\\
Negative Results and Failure Modes in\\
Ensemble AI-Generated Image Detection}

\newcommand{\aut}[2]{%
  \begin{tabular}[t]{c}
    #1 \\
    {\small\texttt{#2}}
  \end{tabular}}

\author{%
  \aut{Suraj Singh}{suraj.singh.nsut@gmail.com}
  \hspace{1.5em}
  \aut{Tushar Verma}{tusharverma.research@gmail.com}
  \hspace{1.5em}
  \aut{Pragyan Singh}{singhpragyan21@gmail.com}
  \\[1.2em]
  \aut{Shaurya Bhav}{shauryabhav@gmail.com}
  \hspace{1.5em}
  \aut{Shivam Kumar}{shivamparashar222@gmail.com}
}
\date{\today}

\begin{document}
\maketitle

\begin{abstract}
We built an ordinary stacking ensemble for AI-generated image detection --- three open
detectors producing five scores, fused by a gradient-boosted meta-learner that treats a
detector's failure as missing data --- deployed it, and then evaluated it against three
controls it should have faced first. This paper reports what the controls found,
including where they overturned our own earlier conclusions.

\textbf{Fusion is worth its cost only when refitted on the target domain.} The shipped
meta-learner, fitted on a separate corpus, does not beat its best single member on
$2000$ StyleGAN faces (AUC $0.9896$ vs $0.9961$; McNemar $p = 1.000$). But a stacker
refitted in-domain beats that member plus a post-hoc calibrator
($\Delta\mathrm{AUC} = +0.0025$, $[+0.0014, +0.0038]$; $p = 3.4\times10^{-4}$). An
earlier draft claimed the calibrated single detector won outright; that comparison mixed
regimes and we correct it here.

\textbf{One corpus is not an evaluation.} On $80$ screenshots every model's AUC interval
contains $0.5$. We can say nothing stronger: the difference between the ensemble's drop
and its best member's is $[-0.185, +0.115]$.

\textbf{Abstention is real, correlated, and mishandled.} With four of five detectors
silent and the survivor reporting ``real'', the system returns $P(\text{AI}) = 0.9985$,
because an all-\texttt{NaN} input scores $0.9995$ in a learner never fitted with
missingness. The three AIDE checkpoints fail together, sharing one preprocessing path. A
quorum rule requiring two distinct architectures prevents both failures with no
retraining.

We also find Corpus A carries a class-conditional JPEG bias severe enough to separate the
classes from the header alone, which limits every in-domain number we report. Code,
harness, hash-identified artifacts and all corrections are released.

\end{abstract}

\section{Introduction}
\label{sec:intro}
Generative image models have made synthetic imagery cheap, photorealistic, and
ubiquitous. The defensive response has split into two families. \emph{Provenance}
approaches --- C2PA Content Credentials~\cite{c2pa} and pixel watermarks such as
SynthID~\cite{synthid2025} --- attach a signal at generation time. They are precise
when present, but they are opt-in, they cover only cooperating generators, and the
signal can be stripped by a screenshot, a re-encode, or a crop.
\emph{Passive detection} asks a model to judge an image with no cooperation from
whoever made it. It is the only option for the images that matter most in practice:
the ones already in circulation, from an unknown generator, with the metadata gone.

Passive detection has a well-documented weakness. Detectors that score near-perfectly
in-domain collapse when the generator changes~\cite{wang2020cnn,ojha2023universal}, and
a benchmark built specifically from images that fooled human annotators found that
almost all off-the-shelf detectors classified them as real~\cite{yan2025aide}. The
field's response has largely been to build better single detectors.

We took the position that field takes: given that every individual detector is
unreliable off its training distribution, combine several of them. We built
\textsc{Quorum}, a stacking ensemble~\cite{wolpert1992stacked} of three heterogeneous
open detectors producing five scores, fused by a gradient-boosted
meta-learner~\cite{chen2016xgboost} that treats a detector's failure as \emph{missing
data} rather than as evidence, and deployed it.

This paper is not a proposal of that system. It is a report of what happened when we
evaluated it against the controls it should have been evaluated against from the start:
a calibrated single member, a second corpus, and abstentions that actually occur rather
than ones injected at random. It failed all three tests. We report this because the
system is unremarkable --- it is what a competent team builds --- and because the
controls that expose it are cheap, general, and largely absent from the literature it
came from.

The abstention design was our centre of gravity, and it is where we were most wrong. In
a deployed pipeline an expert routinely fails to score an image: a decode error, an
unsupported colour profile, an accelerator kernel that silently returns
\texttt{NaN}. The tempting implementation --- return $0.0$ and move on --- is
catastrophic, because $0.0$ in a detector's output space does not mean ``no opinion'';
it means ``confidently real''. A failure is thereby laundered into a confident vote
for the negative class, invisibly, in exactly the situations where the system is
already degraded. We treat abstention as a first-class outcome, propagate it as
missingness~\cite{rubin1976inference} into a learner that handles missingness
natively, and surface it in the verdict.

Three things determine whether such a system is usable and are largely unmeasured in
the literature it comes from. \emph{Abstention}: benchmarks assume every detector
returns a score for every image, while deployed detectors meet truncated files,
unsupported colour spaces and degenerate geometry, and the usual handling --- a
\texttt{try/except} returning $0.0$ --- silently converts an infrastructure failure into
a confident vote for ``real''. \emph{Calibration}: detectors are reported at a fixed
$0.5$ threshold, which off-distribution confounds whether a detector \emph{orders} real
below fake with whether its scores fall on the right side of one cut. \emph{Silent
system failure}: detection papers report model error, not system error, and we
encountered defects that produced confidently wrong verdicts with no exception raised
and no anomaly in the output distribution. Our evaluation targets all three.

Our contributions are:

\begin{enumerate}\itemsep2pt
  \item \textbf{Fusion must be compared like with like, and then it wins.} Zero-shot,
  the shipped stacker ties its best member. Refitted in-domain it beats that member plus
  a calibrator, with a paired interval excluding zero. Our earlier claim that a
  calibrated single detector won outright came from ranking an in-domain calibrator
  against a zero-shot ensemble; \S\ref{sec:results} records the error and the correction.
  The surviving lesson is about \emph{deploying a stacker fitted on a foreign corpus},
  not about stacking.
  \item \textbf{Neither corpus alone would have been informative.} Corpus A reports
  $94.7\%$; on Corpus B every interval contains chance. We do not claim fusion degrades
  worse than its members --- the paired interval $[-0.185, +0.115]$ cannot resolve it ---
  and we identify a compression confound that offers a competing explanation for both.
  \item \textbf{Abstention is real, correlated, and the learner handles it backwards.}
  Four of five detectors silent and one survivor saying ``real'' yields
  $P(\text{AI}) = 0.9985$; an all-\texttt{NaN} vector scores $0.9995$. Abstentions
  cluster by shared preprocessing, so MCAR injection is optimistic. A quorum rule fixes
  both observed cases without retraining --- which is the correction to our own design.
  \item \textbf{A failure catalogue} (\S\ref{sec:silent}): eleven defects that corrupted
  verdicts silently, four of which bear on abstention and seven of which are ordinary
  engineering faults. An experience report from one codebase, not a survey.
\end{enumerate}

We are explicit about scope. Our two corpora cover faces (StyleGAN, $n=2000$) and
screenshots (mixed provenance, $n=80$); neither contains a controlled sample of modern
diffusion output, and the second is small enough that its confidence intervals are wide.
We make no claim of state-of-the-art detection and run no comparison against published
detectors on their own benchmarks. What the second corpus does establish --- because the
collapse is to chance and not to a slightly worse number --- is that conclusions drawn
from either corpus alone would have been wrong. Section~\ref{sec:discussion} states
plainly what this does and does not support.

\section{Related Work}
\label{sec:related}
\textbf{Passive detectors.} Wang et al.~\cite{wang2020cnn} showed a classifier trained
on a single generator transfers across CNN generators via shared upsampling artifacts.
Ojha et al.~\cite{ojha2023universal} identified the failure our results repeatedly
exhibit: a detector trained for real-vs-fake becomes asymmetrically tuned and the real
class turns into a \emph{sink} absorbing anything unfamiliar, including unseen
generators. AIDE~\cite{yan2025aide} combines a low-level pathway over SRM steganalysis
residuals~\cite{fridrich2012rich} with a frozen CLIP-family semantic trunk; it is one of
the three detectors we fuse, and its Chameleon benchmark demonstrated that almost all
off-the-shelf detectors classify hard AI images as real.

\textbf{Detector fusion.} Combining detectors is not new, and two results from that
literature predict findings we report. Kittler et al.~\cite{kittler1998combining} analyse
fixed combination rules and show the $\max$ rule is the most sensitive to a single
over-confident member --- which is what \S\ref{sec:results} rediscovers empirically.
Kuncheva and Whitaker~\cite{kuncheva2003diversity} formalise the diversity--accuracy
relationship and the pairwise-disagreement measure we use. Our \S\ref{sec:results}
fusion-rule results should be read as a replication of this behaviour in a
heterogeneous open-detector stack, not as new. Our \S\ref{sec:intro} characterisation of
the field as building better single detectors describes the dominant line, not the only
one.

\textbf{Compression bias.} JPEG history is a documented confound in this area: Grommelt
et al.~\cite{grommelt2024jpeg} show detection datasets where real and synthetic classes
differ in compression history, so detectors learn compression rather than generation.
\S\ref{subsec:jpeg} finds exactly this in our Corpus A, and our compression error
correlate is a confirmation of that line rather than a new finding.

\textbf{Coincident failure.} That independently built components fail together far more
than independence predicts is the central result of the N-version programming
literature~\cite{knight1986independence}. Our correlated-abstention observation
(\S\ref{sec:results}) is that effect in a detector ensemble, reached empirically.

\textbf{Benchmarks.} GenImage~\cite{zhu2023genimage} provides over a million real/fake
pairs across seven modern generators with explicit cross-generator protocols. It is the
evaluation this work most needs and does not have (\S\ref{sec:discussion}).

\textbf{Provenance.} C2PA Content Credentials~\cite{c2pa} and pixel watermarks such as
SynthID~\cite{synthid2025} attach a signal at generation time. They are precise when
present but cover only cooperating generators, and metadata is routinely stripped in
ordinary distribution~\cite{passiveorwatermark2024}. Passive detection remains the only
option for the images that motivate the problem.

\textbf{What we do not do.} We evaluate no published detector on its own benchmark, and
report no number comparable to a published one. Our corpora are neither
AIGCDetectBenchmark, GenImage, nor Chameleon, and the corpus our fusion model was fitted
on no longer exists. Every comparison in this paper is therefore \emph{internal} ---
fusion rules, detector subsets, calibrators and missingness policies measured against one
another under one protocol. That supports relative claims and supports none about
standing relative to published detectors. Commercial APIs (Hive, Reality Defender, AI or
Not, and others) are excluded for the same reason plus a stronger one: their models and
training data are undisclosed, so a score from them cannot be audited or
reproduced~\cite{fitforpurpose2025}. Where we call a control under-reported we mean
\emph{to our knowledge, for detector ensembles}; we have run no survey, and
second-corpus evaluation in particular is the field's standard cross-generator protocol
rather than a gap.

\section{The System}
\label{sec:method}
\textsc{Quorum} is a two-stage stacked generaliser~\cite{wolpert1992stacked}. Stage one
runs three heterogeneous detectors that emit five scores; stage two fuses those scores
with a gradient-boosted tree ensemble~\cite{chen2016xgboost}. The design commitments
that distinguish it from a textbook stack are the abstention contract
(\S\ref{subsec:abstention}) and the single-authority rule (\S\ref{subsec:fusion}).

\subsection{Detectors and scores}
\label{subsec:experts}

A terminological point, because it bears on how much diversity this ensemble actually
has. \textsc{Quorum} runs \textbf{three detectors} which emit \textbf{five scores}:
AIDE contributes three, one per checkpoint, over a single shared architecture. We use
\emph{detector} for the three models and \emph{score} (or feature) for the five
columns, and we do not claim five independent experts. Section~\ref{sec:results}
quantifies the consequence: the two AIDE checkpoints that share an architecture are the
\emph{least} diverse pair in the system, and architectural rather than checkpoint
diversity is what produces independent errors.

Detectors were chosen for \emph{inductive diversity} rather than individual
strength~\cite{dietterich2000ensemble}: a semantic classifier, a
self-supervised-feature classifier, and a frequency-forensic model, so that their
errors have some chance of being uncorrelated. Each maps an image to
$[0,1]$, interpreted as $P(\text{AI-generated})$.

\begin{table}[h]\centering\small
\setlength{\tabcolsep}{4pt}
\caption{The three third-party detectors and the five scores they produce. None is our
work; we integrate and evaluate them.}
\label{tab:detectors}
\begin{tabular}{lll}
\toprule
Detector & Basis & Scores \\
\midrule
Swin-B~\cite{liu2021swin}   & semantic classifier          & \texttt{VQGAN} \\
Nonescape                   & frozen DINOv2~\cite{oquab2024dinov2} & \texttt{NONESCAPE} \\
                            & + EfficientNetV2 attention   & \\
AIDE~\cite{yan2025aide}     & SRM residuals~\cite{fridrich2012rich} & \texttt{SD14}, \texttt{PROGAN}, \\
                            & over DCT-selected patches    & \texttt{GENIMAGE} \\
\bottomrule
\end{tabular}
\end{table}

\textbf{Provenance.} The Swin classifier is the \texttt{umm-maybe/AI-image-detector}
checkpoint (AutoTrain-produced, CC-BY-4.0; weight digest
\texttt{f4c14ed23eb1d65b1c6ca7b163f8f91b}\ldots). Its own model card states it is ``a
proof-of-concept demonstration'' intended for imagery from older generators such as
VQGAN+CLIP, and directs users to a newer detector instead. We report this because it
bears directly on our results: the detector we find weakest (AUC $0.6189$ on Corpus A,
$0.4817$ on Corpus B) is one its authors had already deprecated, and our recommendation
to drop it is a confirmation of their guidance rather than a new finding. Nonescape is
the v0 release from \texttt{nonescape.com} (Apache-2.0, \texttt{github.com/nonescape},
weights \texttt{nonescape-v0.safetensors}, digest
\texttt{78fd1713c8ceca1fe8caf0cacca61d80}\ldots). AIDE uses the three checkpoints
released with~\cite{yan2025aide}. We redistribute none of these weights.

Three properties of these models matter for our results and are noted here rather than
re-derived later. The Swin classifier's label order is \emph{inverted} relative to the
others, which caused defect S4. AIDE selects the two flattest and two busiest
$32{\times}32$ patches through a DCT module before its forensic pathway --- a step that
fails on degenerate geometry and, because all three checkpoints share it, makes their
abstentions perfectly correlated (\S\ref{sec:results}). Nonescape re-encodes every input
as JPEG at quality 100, a forensic control against compression history acting as a
spurious cue.

\subsection{Missingness-aware fusion}
\label{subsec:fusion}

Let $s_i \in [0,1] \cup \{\bot\}$ be expert $i$'s output, where $\bot$ denotes
abstention. The meta-learner $g$ receives the vector
$\mathbf{s} = (s_1,\dots,s_5)$ with each $\bot$ mapped to \texttt{NaN}, and returns
$P(\text{AI}) = g(\mathbf{s})$.

Gradient-boosted trees are chosen specifically because they handle missingness
natively: at each split, samples with a missing value are routed down a
\emph{learned} default direction, fitted from the training data rather than imposed by
an imputation rule. The learner therefore represents ``expert $i$ said nothing'' as its
own condition, distinct from any value $s_i$ could have taken. This is the mechanism
by which abstention avoids becoming evidence.

The production model is deliberately shallow --- $100$ trees, depth $3$, learning rate
$0.1$ --- because it fuses five already-strong scores; depth here buys overfitting
rather than accuracy.

\textbf{Single authority.} The meta-learner is the only component that emits a verdict.
There is no threshold vote, no $\max$ rule, and no fallback path. An earlier version of
this system had two entry points using \emph{different} fusion rules (stacking in
batch, $\max$ interactively); they could disagree on the same image, and the $\max$
rule made the ensemble exactly as precise as its most trigger-happy member. We report
that rule as a baseline in Section~\ref{sec:results} rather than shipping it.

\subsection{The abstention contract}
\label{subsec:abstention}

Every expert implements
$\texttt{score}(x) \rightarrow \{\text{feature} \mapsto \text{float} \mid \texttt{None}\}$.
Three rules are enforced throughout, each with regression tests:

\begin{enumerate}\itemsep2pt
  \item \textbf{$0.0$ is never a failure sentinel.} A failure returns \texttt{None},
  which becomes \texttt{NaN}. Writing $0.0$ would assert maximal confidence in
  ``real''.
  \item \textbf{Non-finite values are failures, not scores.} Any \texttt{NaN} or
  $\pm\infty$ reaching the fusion layer is converted to \texttt{None} and
  \emph{recorded}. Left in place it would reach the learner as the same missing
  feature but with no failure logged --- a two-expert verdict presented as a
  five-expert consensus. This check exists because that is precisely what happened
  (Section~\ref{sec:silent}).
  \item \textbf{Total failure is not a verdict.} If every detector abstains the system
  raises rather than returning a number. Abstention propagates to the user as a
  \emph{degraded} flag naming which detectors were silent, following the classical
  reject-option framing~\cite{chow1970optimum}.
\end{enumerate}

\textbf{The gap, stated plainly.} Rule~3 defines a safe state for $5/5$ abstention and
for nothing else. At $1$--$4$ abstentions the system emits a live probability computed
from the learner's unfitted defaults, carrying a \texttt{degraded} flag. That is neither
fail-safe nor fail-operational: it is a confident answer with a warning label, and
\S\ref{sec:results} shows the answer can be inverted. The system is named for a quorum it
does not implement --- there is no minimum reporting set anywhere in the contract. We
kept the name because the omission is the paper's most concrete finding, and
\S\ref{sec:results} evaluates the rule that closes it.

A note on vocabulary. Chow's reject option~\cite{chow1970optimum} describes a classifier
declining on a datum it finds uncertain, which is informative about the datum. Our
abstentions are decode errors, unsupported colour modes and accelerator faults, which are
informative about the pipeline. None of our detectors implements a reject option, which
is why $2080$ images produced no natural abstention. This paper studies
\emph{failure-induced} missingness; input-driven rejection is a different construct we do
not evaluate.

\subsection{Implementation}

The system is a single Python package with one execution path shared by every entry
point. Weights total ${\sim}16$\,GB and are held resident; per-expert batched inference
amortises the forward pass across images. A PDF is handled by rasterising each page at
$2\times$ native resolution and scoring pages as images, with the document flagged if
any page is. Model loading is lazy, and inference is fully offline after a single
warm-up that populates a repository-local model cache.

\section{Silent Failures: An Experience Report}
\label{sec:silent}
\emph{Scope.} This section is an experience report, not a study. It catalogues defects
encountered while building one system, by one team, on one codebase --- $n = 1$. We
present it because the defects are concrete, reproducible from the descriptions given,
and of a kind that the metrics this field reports cannot detect. We do \emph{not} claim
evidence that they are prevalent elsewhere: establishing that would require surveying
independent implementations, which we have not done. Read what follows as a hypothesis
about a failure class, supported by existence proofs rather than by frequency
estimates.

The common property is that each fault corrupts verdicts while raising no error and
leaving no anomaly in the output distribution.

\paragraph{S1. Accelerator miscompilation (severity: critical).}
On Apple Silicon, the MPS backend of PyTorch 2.7.1 miscompiled AIDE's fused forward
pass, returning all-\texttt{NaN} logits for every image, deterministically, while the
identical computation on CPU returned correct values. The input tensors were finite on
both devices; merely registering a forward hook --- which breaks kernel fusion --- also
restored correct values, indicating a compilation rather than an algorithmic fault.

The consequence is the paper's motivating example. $\mathrm{softmax}(\texttt{NaN})$ is
\texttt{NaN}, so all three AIDE features became missing; the meta-learner routed them
down its learned default branches; and a verdict computed from two experts was reported
as a five-expert consensus with no degradation flag. The same document was observed
receiving $P(\text{AI}) = 1.00$ and $P(\text{AI}) = 0.02$ on different runs.
\emph{Mitigation:} a seeded probe at load time asks the device whether it returns
finite logits for a fixed dummy input; on failure the expert relocates to CPU and logs
it. A future runtime that fixes the bug passes the probe and regains the fast path
with no code change.

\paragraph{S2. $0.0$ as a failure sentinel (critical).}
Every expert originally returned $0.0$ on a decode error. Since $0.0$ means
``confidently real'', unreadable images became confident negative votes --- and,
because the same code produced the training CSVs, this was also baked into the
meta-learner's training data.

\paragraph{S3. Non-finite scores treated as valid (critical).}
Independently of S1, any \texttt{NaN} reaching fusion is indistinguishable from a
genuinely missing feature to the learner, but leaves \texttt{degraded} unset. The
verdict is then presented with more apparent support than it has.

\paragraph{S4. Inverted label polarity (high).}
The Swin expert declares \texttt{id2label = \{0: artificial, 1: human\}}, inverted
relative to every other expert. Reading index $1$ as ``fake'' --- the convention
elsewhere in the system --- silently inverts that expert. \emph{Mitigation:} look the
class up by name and abstain if the expected label is absent.

\paragraph{S5. Divergent fusion rules (high).}
Two entry points used different fusion rules, so the same image could receive different
verdicts depending on which was called.

\paragraph{S6. Filename-derived feature names (high).}
Feature columns were minted by uppercasing CSV filenames, so renaming a file silently
produced a feature the trained model had never seen. \emph{Mitigation:} an explicit
registry mapping each expert to its feature column, shared by training and inference.

\paragraph{S7. Destructive retraining (high).}
The training script overwrote the production model in place with no backup, and
destroyed it once. \emph{Mitigation:} refuse to run without an explicit flag, and take
a timestamped backup unconditionally.

\paragraph{S8. Append-mode CSV accumulation (medium).}
Score CSVs were opened in append mode, so re-running a scoring pass silently doubled
every row, corrupting any model trained on the result.

\paragraph{S9. Filename collision across classes (medium).}
Rows were keyed by bare filename, so \texttt{real/photo.jpg} and \texttt{ai/photo.jpg}
collided --- one silently overwriting the other \emph{with the opposite label}.

\paragraph{S10. Dotfile contamination (medium).}
Directory listings included \texttt{.DS\_Store}, which entered the sample list as a
bogus image and desynchronised filenames from scores.

\paragraph{S11. Device-dependent preprocessing crash (medium).}
AIDE's DCT patch selector indexes with a zero-dimensional tensor, which is valid on CPU
but raises on MPS. Upstream never encounters it because the module always runs in a CPU
dataloader worker.

\paragraph{Common structure.}
Nine of these eleven defects share a shape: \emph{a failure was represented as a value
in the normal output range}. \texttt{NaN}, $0.0$, an inverted label, and a stale
filename are all silently type-correct. The general mitigation is to make failure a
distinct type that cannot be confused with a measurement, and to test that property
directly --- which is what the abstention contract in \S\ref{subsec:abstention}
encodes.

We note, however, that Section~\ref{sec:results} shows this mitigation is necessary but
not sufficient: typing failure correctly still leaves open \emph{what the fusion model
does} with a correctly-typed absence, and our own answer to that turns out not to be the
best one available.

\section{Experimental Setup}
\label{sec:setup}
\subsection{Corpora}

\paragraph{Corpus A --- faces.} The \texttt{whichfaceisreal} corpus: $2000$ images,
balanced $1000$ real / $1000$ synthetic, all $1024{\times}1024$ JPEG. It is distributed
in the \texttt{0\_real}/\texttt{1\_fake} layout of the ForenSynths test set released with
Wang et al.~\cite{wang2020cnn}, and we believe it to be that split; published per-detector
numbers on it therefore exist, and our \S\ref{sec:related} statement that we report no
comparable number should be read as ``we did not run the comparison'', not ``none is
possible''. The synthetic half was produced by
StyleGAN~\cite{karras2019stylegan} and published through the
\texttt{whichfaceisreal.com} perceptual-discrimination study; the real half is drawn
from FFHQ (Flickr-Faces-HQ), the Creative-Commons photographic corpus StyleGAN was
itself trained on.

Two properties of this corpus matter for interpreting our results, and we state both
plainly.

\textbf{It is narrow.} One semantic domain (aligned human faces) and one generator
family (a 2019 GAN). It contains no diffusion imagery. Results here bound
GAN-face performance and say nothing directly about Stable Diffusion or Midjourney.

\textbf{Held-out status: bounded, not established.} The meta-learner was fitted on a
separate corpus of $67{,}231$ images that no longer exists. One artifact of it survives:
a per-image score table listing all $67{,}231$ filenames. Intersecting those filenames
with both corpora gives \textbf{zero overlap} ($0$ of $2000$ for Corpus A, $0$ of $80$
for Corpus B). That is a real bound and better than the ``unverifiable'' we claimed
previously, but it is filename-level only: a renamed or re-encoded duplicate would not be
detected, and FFHQ is common enough in corpora of that period that content overlap cannot
be excluded. The shipped model's learned base score is $0.4999$, indicating a
near-balanced training set. We therefore describe Corpus A as \emph{held out as far as we
can verify}, and use that same wording everywhere.

Because real and synthetic halves share an origin (StyleGAN was trained on FFHQ), face
alignment and framing are not confounded between classes. \textbf{JPEG history,
however, is badly confounded}: measured after review, the real half carries a single
quantisation table with median sum $1483$ and the synthetic half a single different table
with median sum $375$ (\S\ref{subsec:jpeg}). The classes are separable from the JPEG
header alone. This is the dataset bias Grommelt et al.~\cite{grommelt2024jpeg} document,
it limits every in-domain number we report, and we found it only because a reviewer asked.

\paragraph{Corpus B --- screenshots.} $80$ images, $50$ real / $30$ AI, all RGBA PNG at
varied resolutions. Provenance: a third-party collection titled
\texttt{AI\_detection\_dataset}, obtained as a redistributed archive; the curator,
labelling procedure and generator list are not documented by the source beyond filenames
(one is a named Gemini generation). We did not create or verify the labels and cannot
redistribute the images; the per-image score table is released, which supports every
analysis here but does not let a reader inspect the pixels. A reader who needs that must
treat Corpus B as unreproducible. It exists to answer one question Corpus A cannot: does anything here
transfer? It is deliberately different --- a different domain, a different capture path,
and generator provenance that includes at least one diffusion-era model (a Gemini
generation) rather than a 2019 GAN.

We checked the obvious confound before using it. Both classes are screenshots ($50/50$
real, $29/30$ AI), so the classes do not differ by capture method, which would otherwise
make the task trivially separable on an artefact rather than on generation. Three
limitations are unavoidable and we state them rather than discount the corpus: $n = 80$
gives wide intervals, the classes are imbalanced ($5{:}3$), and we did not produce the
labels ourselves, so label provenance rests on the corpus author. Screenshot capture
also applies rescaling and recompression to both classes equally --- which, given our
error analysis (\S\ref{sec:results}), is likely part of why performance falls.

Corpus B is too small to establish a detector's accuracy. It is amply large enough for
the finding we draw from it, which is that every model's confidence interval contains
chance.

\subsection{Protocol}

Scoring is separated from analysis. One pass computes all five expert scores for all
$2000$ images and writes them to a table; every experiment then re-reads that table.
This makes the expensive step happen once and the ablations cheap enough to run
honestly.

We report two regimes and never rank across them:

\begin{description}\itemsep2pt
  \item[Zero-shot] The shipped meta-learner, fitted on the separate corpus above,
  applied with no refitting. This measures cross-domain generalization.
  \item[In-domain CV] Models refitted on this corpus under stratified $5$-fold
  cross-validation, scored strictly out-of-fold. This measures the ceiling available
  here and permits a fair comparison between fusion rules.
\end{description}

Confidence intervals are percentile bootstrap over $2000$ resamples of images (not of
predictions). Paired comparisons use an exact McNemar
test~\cite{mcnemar1947note} on thresholded decisions. Calibration is expected
calibration error over $10$ equal-width confidence bins~\cite{guo2017calibration}.
Unless stated otherwise the decision threshold is $0.5$, matching the deployed rule.
We additionally report each model's Youden-optimal threshold and the accuracy there, to
separate ranking quality from calibration.

\subsection{Abstention experiments}

Neither corpus produced a single natural abstention across $2080$ images, so the
missingness policy cannot be settled observationally. We approach it two ways.

\textbf{Injected (MCAR).} Abstentions are inserted uniformly at random at rates
$\{0, 10, 25, 50\}\%$ over the score matrix, and three policies compared under identical
folds: \textbf{native} (leave \texttt{NaN} for the learner to route),
\textbf{zero-imputed} (substitute $0.0$, the historical bug S2), and
\textbf{mean-imputed} (substitute the column mean). This is a controlled but
unrealistic mechanism: real abstentions are neither uniform nor independent.

\textbf{Induced (real).} We construct $14$ inputs that a deployed system genuinely
receives and that genuinely break preprocessing: CMYK, grayscale, palette and RGBA
colour modes; $16$-bit depth; degenerate geometry ($1{\times}1$, $4096{\times}3$); a
$6000{\times}6000$ image; an animated GIF; a truncated PNG; a corrupt CRC; a zero-byte
file; and prose behind a \texttt{.jpg} extension. Here the abstentions are whatever the
detectors actually do, with whatever correlation structure they actually have.

\subsection{Environment}

Apple M4 Max, $128$\,GB unified memory, macOS 26.6.2, PyTorch 2.7.1, Python 3.11.15.
Following the S1 probe, AIDE executes on CPU; the remaining detectors use the MPS
backend. All randomness is seeded ($42$).

Two selection biases present in an earlier version of this analysis have been removed.
The best single detector is now chosen by AUC \emph{within each training fold}, never on
evaluation data, and every Youden threshold is likewise fitted on training rows and
applied once to held-out rows. We report which detector each fold selected, so that a
selection that is unstable across folds cannot masquerade as a single well-defined
baseline.

\section{Results}
\label{sec:results}
We report Corpus A first because it is where the system looks good, then the three
controls that undo that impression: a second corpus, a calibrated single detector, and
abstentions that actually occur.

\subsection{Corpus A: per-detector and fused}

\begin{table*}[t]\centering\small
\caption{Corpus A ($2000$ StyleGAN faces), zero-shot. \textbf{$t^\star$ and Acc$^\star$
are whole-corpus (oracle) Youden fits}, retained only to expose the ranking/calibration
gap; they are optimistically biased and are \emph{not} the basis of any claim. The
unbiased nested figures are in \S\ref{sec:results} below. No detector abstained.}
\label{tab:experts}
\begin{tabular}{lrrrrrrrr}
\toprule
Model & Acc & Prec & Rec & F1 & AUC & AP & $t^\star$ & Acc$^\star$ \\
\midrule
AIDE-SD14        & 0.7455 & 0.9168 & 0.5400 & 0.6797 & 0.8821 & 0.8842 & 0.154 & 0.7910 \\
VQGAN/Swin       & 0.5695 & 0.6304 & 0.3360 & 0.4384 & 0.6189 & 0.6089 & 0.347 & 0.5895 \\
AIDE-PROGAN      & \textbf{0.9470} & 0.9923 & 0.9010 & 0.9444 & \textbf{0.9961} & \textbf{0.9962} & 0.322 & \textbf{0.9700} \\
AIDE-GENIMAGE    & 0.7420 & 0.9764 & 0.4960 & 0.6578 & 0.9208 & 0.9254 & 0.059 & 0.8535 \\
NONESCAPE        & 0.7960 & 0.7139 & 0.9880 & 0.8289 & 0.9567 & 0.9520 & 0.881 & 0.8965 \\
\midrule
\textsc{Quorum} (shipped) & \textbf{0.9470} & 0.9357 & 0.9600 & \textbf{0.9477} & 0.9896 & 0.9899 & 0.555 & 0.9515 \\
\bottomrule
\end{tabular}
\end{table*}

\paragraph{Fusion does not beat its best member.}
AIDE-PROGAN alone attains AUC $0.9961$ against the ensemble's $0.9896$. At $t=0.5$ the
two are identical to four decimals, and an exact McNemar test finds them
indistinguishable: $81$ images correct only for the single detector, $81$ only for the
ensemble, $p = 1.000$ (Holm-adjusted $1.000$).

An earlier version of this analysis selected that detector by AUC over the whole corpus
--- an oracle no deployment has --- and fitted each Youden threshold on the same rows it
scored. Both biases are removed by nesting: the detector and every threshold are now
chosen inside training folds and applied once to held-out rows. The conclusion survives
and strengthens. All five folds selected AIDE-PROGAN \emph{unanimously}, so the baseline
is a stable object rather than an artefact of one split, and under nested thresholds the
single detector leads on accuracy, $0.9640$ against the ensemble's $0.9510$. The
per-fold thresholds were $\{0.3217, 0.3217, 0.3882, 0.3217, 0.2698\}$ for the detector
and $\{0.5547, 0.5547, 0.5547, 0.5547, 0.5462\}$ for the ensemble; both are stable, so
the nested figures are not averaging over wildly different operating points. Note that
this comparison is zero-shot on both sides --- it is \emph{not} the in-domain comparison
of \S\ref{sec:results}, where fusion wins.

The explanation is mundane. AIDE-PROGAN is the checkpoint trained on ProGAN, a GAN; the
corpus is StyleGAN, a GAN. This is the one configuration where a specialist's training
distribution matches the test generator almost exactly, and fusion cannot exceed a
member that is already at ceiling.

\begin{figure}[t]\centering
\includegraphics[width=\linewidth]{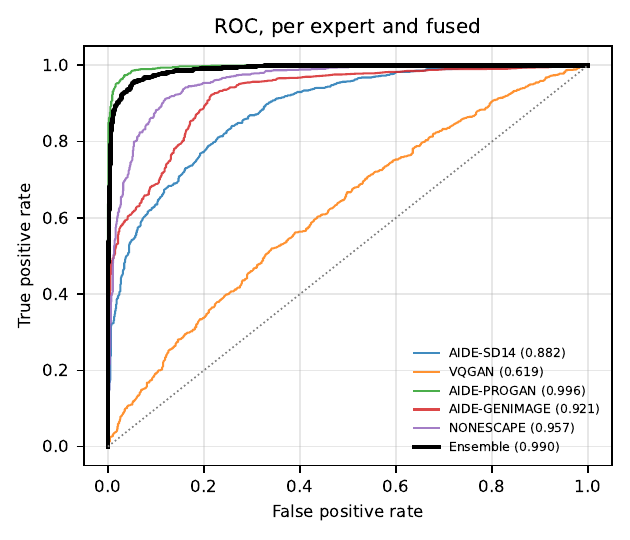}
\caption{Corpus A ROC. AIDE-PROGAN (0.9961) sits above the fused curve (0.9896).}
\label{fig:roc}
\end{figure}

\begin{figure}[t]\centering
\includegraphics[width=\linewidth]{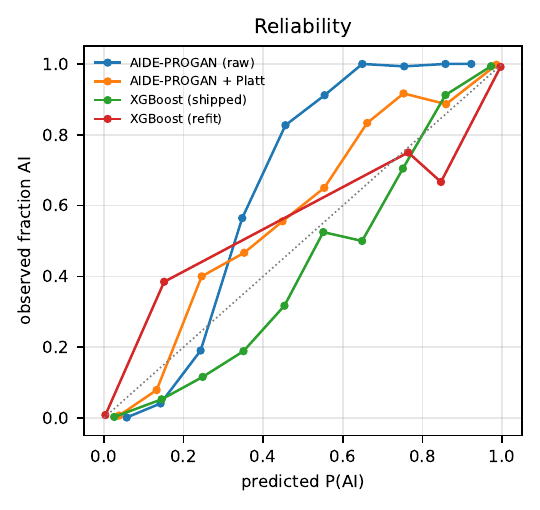}
\caption{Corpus A reliability, including the single detector raw and Platt-scaled.}
\label{fig:reliability}
\end{figure}

\subsection{Control 1: a calibrated single detector}

The natural retreat from the result above is that fusion still emits a well-calibrated
probability while a raw detector score is not one. Testing that retreat exposed an error
in an earlier version of this paper, which we report rather than quietly repair.

\begin{table*}[t]\centering\small
\caption{Corpus A. Calibrators are fitted strictly out-of-fold \emph{on Corpus A},
so they are in-domain models and belong in the lower block. ECE intervals are
bootstrap over 2000 image resamples.}
\label{tab:calibrated}
\begin{tabular}{llrrl}
\toprule
Model & Regime & AUC & Acc & ECE [95\% CI] \\
\midrule
AIDE-PROGAN (raw)         & zero-shot & 0.9961 & 0.9470 & 0.1272 [0.1195, 0.1349] \\
\textsc{Quorum} (shipped) & zero-shot & 0.9896 & 0.9470 & 0.0416 [0.0348, 0.0512] \\
\midrule
AIDE-PROGAN + Platt       & in-domain & 0.9960 & 0.9660 & 0.0319 [0.0272, 0.0392] \\
AIDE-PROGAN + isotonic    & in-domain & 0.9946 & 0.9665 & \textbf{0.0083} [0.0074, 0.0175] \\
XGBoost (refit)           & in-domain & \textbf{0.9985} & \textbf{0.9805} & 0.0096 [0.0063, 0.0163] \\
\bottomrule
\end{tabular}
\end{table*}

\paragraph{A correction.} An earlier draft reported that ``two Platt parameters on one
detector beat the whole ensemble on AUC, accuracy and calibration at once''. That
comparison put an \emph{in-domain} calibrator, fitted out-of-fold on Corpus A labels,
against the \emph{zero-shot} shipped artifact --- exactly the cross-regime ranking
\S\ref{sec:setup} forbids. The claim was an artifact of the regime mixing. We state the
corrected result in both regimes.

\textbf{Zero-shot.} The shipped ensemble does not beat the raw best member: AUC $0.9896$
against $0.9961$, accuracy tied at $0.9470$, McNemar $p = 1.000$. A calibrated single
detector has no zero-shot counterpart, because fitting the calibrator requires target
labels. This half of the original claim survives.

\textbf{In-domain.} Given the same Corpus A labels the calibrator was given, fusion
wins, and the margin is resolvable. Against Platt, refit fusion gains
$\Delta\mathrm{AUC} = +0.0025$ $[+0.0014, +0.0038]$ (paired bootstrap, excludes zero)
and $0.9805$ vs $0.9660$ accuracy, McNemar $46$ vs $17$ discordant, $p = 3.4\times
10^{-4}$. Against isotonic, $\Delta\mathrm{AUC} = +0.0039$ $[+0.0022, +0.0060]$,
$p = 2.3\times10^{-4}$. Against the raw best member, $+0.0024$ $[+0.0013, +0.0037]$.
On calibration the two are \emph{not} separable: isotonic's ECE interval
$[0.0074, 0.0175]$ overlaps refit fusion's $[0.0063, 0.0163]$.

\textbf{What this means.} Our headline negative claim about fusion does not survive its
own control. Stated correctly: \emph{a stacker frozen from a foreign corpus does not beat
its best member; a stacker refitted on the target domain does, including against that
member plus a calibrator.} The practical reading is not ``do not fuse'' but ``do not
deploy a stacker fitted elsewhere, and compare like with like''. We record the error
because the regime-mixed version is the more quotable claim and we published it to
ourselves for three revisions before the control caught it.

\subsection{Control 2: a second corpus}

\begin{table}[t]\centering\small
\caption{The same fixed models on both corpora. Corpus B intervals are percentile
bootstrap over $2000$ resamples. Every Corpus B interval contains $0.5$.}
\label{tab:corpusb}
\setlength{\tabcolsep}{4pt}
\begin{tabular}{lrrl}
\toprule
Model & A (AUC) & B (AUC) & B 95\% CI \\
\midrule
AIDE-SD14      & 0.8821 & 0.6293 & [0.4993, 0.7393] \\
VQGAN/Swin     & 0.6189 & 0.4817 & [0.3462, 0.6119] \\
AIDE-PROGAN    & 0.9961 & 0.4563 & [0.3311, 0.5857] \\
AIDE-GENIMAGE  & 0.9208 & 0.4227 & [0.2894, 0.5584] \\
NONESCAPE      & 0.9567 & 0.4620 & [0.3289, 0.5958] \\
\midrule
\textsc{Quorum} & 0.9896 & 0.4107 & [0.2730, 0.5474] \\
\bottomrule
\end{tabular}
\end{table}

On Corpus B the ensemble scores AUC $0.4107$ and accuracy $0.5000$ --- chance, on a
balanced-enough corpus, from a system that scored $94.7\%$ on Corpus A. Every model's
confidence interval contains $0.5$. We therefore claim only that nothing here is
distinguishable from chance; the point estimates sit below $0.5$, but at $n = 80$ we
cannot assert that they are genuinely inverted.

Two things about this table matter more than the individual numbers.

\textbf{The best detector on A is among the worst on B.} AIDE-PROGAN falls from
$0.9961$ to $0.4563$, a drop of $0.5397$. Its dominance on Corpus A was not a property
of the detector but of the coincidence between its training generator and that corpus's.

\textbf{The ensemble's point drop is the largest, but that ordering is not
resolvable.} The ensemble falls $0.5789$ and the best member $0.5397$, a difference of
$0.0391$. A paired bootstrap on the difference of drops gives $[-0.1849, +0.1151]$,
which contains zero. An earlier draft read this ordering as ``fusion amplified the
distribution shift rather than hedging against it''. That claim is not supported and we
have withdrawn it: at $n = 80$ the two drops cannot be distinguished. What survives is
that fusion did not visibly \emph{protect} against the shift either.

\textbf{Baselines, stated numerically.} ``Chance'' means AUC $= 0.5$ and, for accuracy on
this $50/30$ corpus, the majority-class rate of $0.625$. The ensemble's $0.5000$ accuracy
is below the majority-class baseline, and its AUC interval contains $0.5$. We avoid
``collapses to chance'' as a bare assertion: AIDE-SD14's interval $[0.4993, 0.7393]$
touches $0.5$ only at its lower edge and is equally consistent with AUC $\approx 0.63$.

\subsection{A confound in Corpus A}
\label{subsec:jpeg}

Prompted by review, we measured what \S\ref{sec:setup} had only flagged as possible. The
two classes of Corpus A have systematically different JPEG histories
(Table~\ref{tab:jpeg}).

\begin{table}[t]\centering\small
\caption{Corpus A JPEG encoding by class, first 250 images each. Lower quantisation sums
mean higher quality. Each class carries exactly \emph{one} distinct quantisation table.}
\label{tab:jpeg}
\setlength{\tabcolsep}{4pt}
\begin{tabular}{lrr}
\toprule
 & real (FFHQ) & AI (StyleGAN) \\
\midrule
median quant.-table sum & 1483 & 375 \\
median bytes/pixel      & 0.2399 & 0.4253 \\
distinct quant. tables  & 1 & 1 \\
\bottomrule
\end{tabular}
\end{table}

This is the class-conditional compression bias Grommelt et
al.~\cite{grommelt2024jpeg} document for generated-image detection datasets, and it is
severe here: every real image shares one quantisation table and every synthetic image
shares a different one, so the two classes are separable from the JPEG header alone,
without reading a single pixel. We did not run this measurement before the main
experiments, and it materially weakens Corpus A.

Three consequences follow, and we state them rather than discount them.

\textbf{Corpus A separability is partly a corpus artifact.} We cannot say how much of
AIDE-PROGAN's $0.9961$ AUC, or of any in-domain number, rests on generation artifacts
versus compression history. Three of five detectors consume SRM residuals, which is
precisely the signal JPEG quantisation alters.

\textbf{The \S\ref{sec:results} compression correlate is confounded by construction.}
That misclassified images are more compressed ($0.2574$ vs $0.3589$ bytes/pixel) is a
real measurement, but on a corpus where compression already tracks the label it cannot
be read as a clean statement about detector inductive bias.

\textbf{Compression is a live alternative explanation for Corpus B.} Corpus B is
screenshots --- uniformly rescaled and recompressed, which destroys the class-conditional
JPEG signature that Corpus A supplies. The collapse in
Table~\ref{tab:corpusb} is therefore equally consistent with ``these detectors lose a
compression cue they were relying on'' as with ``these detectors fail to transfer across
domains''. Separating the two requires re-encoding both Corpus A classes identically and
re-scoring, which we have not done. Until then Control 2 should be read as a
\emph{transfer-or-compression} failure, not as evidence about domain shift specifically.

\subsection{Control 3: abstentions that actually occur}

Across $2080$ images in both corpora, \emph{not one detector abstained even once}. The
construct this system was designed around never appeared in either evaluation. So we
induced it (Table~\ref{tab:induced}).

\begin{table}[t]\centering\small
\caption{Induced failures. $14$ constructed inputs; the table lists the outcomes that
were not "scored, nothing abstained".}
\label{tab:induced}
\setlength{\tabcolsep}{4pt}
\begin{tabular}{llrr}
\toprule
Input & Outcome & Abst. & $P(\text{AI})$ \\
\midrule
baseline RGB      & scored               & 0/5 & 0.0313 \\
$1{\times}1$ px   & scored, degraded     & 4/5 & \textbf{0.9985} \\
$4096{\times}3$   & scored, degraded     & 3/5 & \textbf{0.9991} \\
prose as .jpg     & all detectors failed & 5/5 & --- \\
zero bytes        & all detectors failed & 5/5 & --- \\
corrupt CRC       & all detectors failed & 5/5 & --- \\
\bottomrule
\end{tabular}
\end{table}

Colour-mode and depth pathologies --- CMYK, grayscale, palette, RGBA, $16$-bit --- were
all handled without abstention, as were a $6000{\times}6000$ image, an animated GIF, and
a truncated PNG. Genuinely unreadable inputs correctly raised rather than returning a
verdict. No case produced an uncaught exception. To that extent the abstention contract
(\S\ref{subsec:abstention}) works as specified.

The two partial cases are where it fails, and they fail badly. On the $1{\times}1$ image
four of five detectors abstain; the survivor, NONESCAPE, reports $0.019$ --- confidently
"real" --- and the fusion returns $P(\text{AI}) = 0.9985$. On the $4096{\times}3$ sliver
three abstain; both survivors report $0.2824$ and $0.2245$, again "real", and the fusion
returns $0.9991$. In both cases every piece of live evidence points one way and the
system reports near-certainty in the other. The \texttt{degraded} flag does fire, which
is the difference between a wrong answer and an unmarked wrong answer, but the
probability itself is not merely unreliable --- it is inverted.

\textbf{Abstentions are correlated, not independent.} All three AIDE checkpoints abstain
together in both cases, because they share one DCT preprocessing path: a geometric
degeneracy that breaks patch extraction breaks it for all three at once. Our injected
experiment sampled missingness independently per score, which cannot produce this and is
therefore optimistic about exactly the situation that occurs.

\paragraph{Both failures are prevented by a rule outside the learner.}
The system is named for a quorum it does not implement: \S\ref{subsec:abstention}'s
rule~3 defines a safe state only for $5/5$ abstention, so $4/5$ and $3/5$ produce a live
verdict. Prompted by review we scored the two cases under two supervisory rules that
require no retraining (Table~\ref{tab:quorum}).

\begin{table}[t]\centering\small
\setlength{\tabcolsep}{4pt}
\caption{Supervisory rules applied to the two partial-failure cases. Quorum: abstain
unless at least two \emph{distinct architectures} report. Survivor-mean: average the
live scores instead of consulting the learner's defaults.}
\label{tab:quorum}
\scriptsize
\begin{tabular}{lrlr}
\toprule
Case & shipped $P$ & live architectures & rule output \\
\midrule
$1{\times}1$ px & 0.9985 & nonescape & \textbf{abstain} \\
$4096{\times}3$ & 0.9991 & nonescape, swin & \textbf{0.2535} \\
\bottomrule
\end{tabular}
\end{table}

Both inversions disappear. On the $1{\times}1$ image only one architecture survives, so
the quorum rule abstains rather than emitting $0.9985$. On the sliver two architectures
survive and the survivor-mean returns $0.2535$ --- ``real'', agreeing with both live
detectors, against the learner's $0.9991$. Neither rule requires the learner to have
seen the failure: both are fixed, inspectable, and evaluated here at no compute cost.
This is the engineered alternative to training on missingness, and
\S\ref{sec:discussion} argues it is the one we should have shipped.

\subsection{Why: what a substituted value asserts}

\begin{table}[t]\centering\small
\caption{Corpus A prior is $0.5000$; a policy meaning "no evidence" should land near it.}
\label{tab:mechanism}
\begin{tabular}{lrr}
\toprule
Substitute & mean $P(\text{AI})$ & $|$dist. from prior$|$ \\
\midrule
all-zero        & 0.0006 & 0.4994 \\
all-column-mean & 0.5990 & \textbf{0.0990} \\
all-\texttt{NaN}  & 0.9995 & 0.4995 \\
\bottomrule
\end{tabular}
\end{table}

Feeding the model a vector consisting entirely of one substitute explains
Table~\ref{tab:induced} exactly. An all-zero vector yields $P(\text{AI}) = 0.0006$, the
expected assertion of "certainly real" that motivated our design. But an
all-\texttt{NaN} vector yields $0.9995$: "certainly AI". \emph{The two are almost
exactly equally extreme}, $0.4994$ and $0.4995$ from the prior, differing only in
direction.

This contradicts the assumption the system ships on. Propagating \texttt{NaN} into a
gradient-boosted model is only a representation of absent evidence if the model was
fitted with absence present; ours was not, so its default branch directions were never
estimated for that purpose and here point almost unanimously toward the positive class.
Mean-imputation wins because the column mean is the only substitute approximating the
marginal, landing $0.0990$ from the prior instead of $0.4994$.

The defensible conclusion is narrower than the abstention contract implies. What matters
is that the fusion model be \emph{trained with missingness represented as it will
occur}, including its correlation structure. Absent that, \texttt{NaN} is not a safe
default but a confident assertion in whichever direction unfitted defaults happen to
point. Its one real advantage over $0.0$ is operational, not statistical: it carries a
\texttt{degraded} flag, so the assertion is visible.

\subsection{Injected missingness, for comparison}

\begin{table}[t]\centering\small
\caption{Corpus A. Fixed model trained on complete scores; abstentions injected MCAR at
inference.}
\label{tab:missing}
\begin{tabular}{llrrr}
\toprule
Rate & Policy & AUC & Acc & mean $P(\text{AI})$ \\
\midrule
0\%  & any           & 0.9985 & 0.9805 & 0.4994 \\
\midrule
10\% & native (NaN)  & 0.9881 & 0.9375 & 0.5462 \\
10\% & zero-imputed  & 0.9824 & 0.9325 & 0.4472 \\
10\% & mean-imputed  & 0.9953 & 0.9605 & 0.5120 \\
\midrule
25\% & native (NaN)  & 0.9580 & 0.8625 & 0.6226 \\
25\% & zero-imputed  & 0.9511 & 0.8390 & 0.3497 \\
25\% & mean-imputed  & 0.9827 & 0.9220 & 0.5304 \\
\midrule
50\% & native (NaN)  & 0.8780 & 0.7330 & 0.7557 \\
50\% & zero-imputed  & 0.8727 & 0.6965 & \textbf{0.2056} \\
50\% & mean-imputed  & \textbf{0.9377} & \textbf{0.8370} & 0.5551 \\
\bottomrule
\end{tabular}
\end{table}

\begin{table}[t]\centering\small
\caption{Corpus A. Model \emph{refitted} under each policy at each rate --- the
like-for-like comparison. Table~\ref{tab:missing} characterises the shipped artifact;
this characterises the design.}
\label{tab:refit}
\begin{tabular}{llrr}
\toprule
Rate & Policy & AUC & Acc \\
\midrule
0\%  & any           & 0.9985 & 0.9805 \\
\midrule
10\% & native (NaN)  & 0.9972 & 0.9720 \\
10\% & zero-imputed  & 0.9970 & 0.9755 \\
10\% & mean-imputed  & 0.9974 & 0.9710 \\
\midrule
25\% & native (NaN)  & 0.9947 & 0.9575 \\
25\% & zero-imputed  & 0.9936 & 0.9595 \\
25\% & mean-imputed  & 0.9946 & 0.9570 \\
\midrule
50\% & native (NaN)  & 0.9722 & 0.9075 \\
50\% & zero-imputed  & 0.9731 & 0.9035 \\
50\% & mean-imputed  & 0.9717 & 0.9000 \\
\bottomrule
\end{tabular}
\end{table}

Under MCAR injection the same ordering appears in milder form: zero-imputation drives
mean $P(\text{AI})$ to $0.2056$ at a $50\%$ rate, \texttt{NaN} drifts to $0.7557$, and
mean-imputation stays nearest calibrated and best on accuracy at every rate. When the
model is instead \emph{refitted} under each policy the policies converge to within
$0.002$ AUC --- a tree ensemble simply learns whatever convention it is shown. The gap
between this table's mild degradation and Table~\ref{tab:induced}'s inverted verdicts is
the cost of assuming independence.

\begin{figure*}[t]\centering
\includegraphics[width=\textwidth]{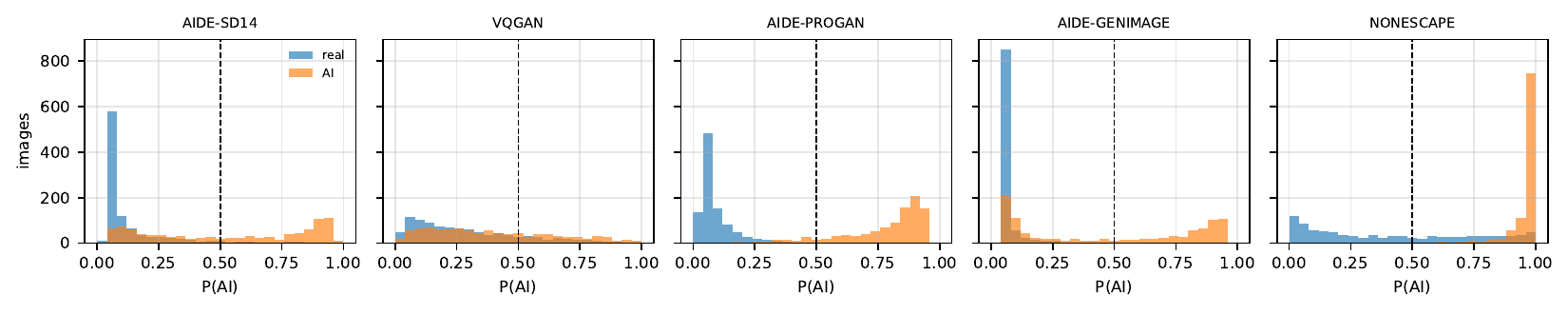}
\caption{Corpus A score distributions per detector, real vs AI, with the $t=0.5$ line.
AIDE-SD14 and AIDE-GENIMAGE compress both classes below the line --- the ``sink class''
effect --- so they appear to fail at $0.5$ while still separating the classes.}
\label{fig:dist}
\end{figure*}

\subsection{Fusion rules and detector redundancy}

\begin{table*}[t]\centering\small
\caption{Corpus A fusion rules. Regimes are never ranked against each other. Brackets
are 95\% bootstrap CIs over 2000 resamples.}
\label{tab:fusion}
\begin{tabular}{llrlrlr}
\toprule
Rule & Regime & AUC & 95\% CI & Acc & 95\% CI & ECE \\
\midrule
Best single (AIDE-PROGAN) & untrained    & 0.9961 & [0.9944, 0.9974] & 0.9470 & [0.9370, 0.9565] & 0.1272 \\
Max fusion                & untrained    & 0.9628 & [0.9548, 0.9702] & 0.7185 & [0.6985, 0.7375] & 0.2589 \\
Mean fusion               & untrained    & 0.9913 & [0.9886, 0.9938] & 0.8710 & [0.8570, 0.8855] & 0.2279 \\
Majority vote             & untrained    & 0.9685 & [0.9626, 0.9742] & 0.8550 & [0.8395, 0.8695] & 0.1544 \\
\textsc{Quorum} (shipped) & zero-shot    & 0.9896 & [0.9861, 0.9925] & 0.9470 & [0.9365, 0.9565] & 0.0416 \\
\midrule
Logistic regression       & in-domain CV & \textbf{0.9989} & [0.9982, 0.9994] & \textbf{0.9815} & [0.9755, 0.9870] & 0.0112 \\
XGBoost (refit)           & in-domain CV & 0.9985 & [0.9977, 0.9992] & 0.9805 & [0.9740, 0.9860] & \textbf{0.0096} \\
\bottomrule
\end{tabular}
\end{table*}

\textbf{The $\max$ rule is catastrophic}: $0.7185$ accuracy, $22.9$ points below learned
fusion ($497$ images correct only for the ensemble against $40$ for $\max$,
$p \approx 2\times10^{-101}$, Holm-adjusted $6\times10^{-101}$). It inherits NONESCAPE's
$396$ false positives, because one over-firing member sets the maximum. This is a
caution against $\max$-style fusion wherever member precision varies.

\textbf{Gradient boosting is not essential.} Logistic regression matches XGBoost
in-domain with overlapping intervals. With five well-behaved inputs there is little
non-linear structure to exploit; XGBoost's role here is its missing-value handling, and
\S\ref{sec:results} has just shown that handling to be a liability rather than an asset
when the model is not fitted with missingness.

\begin{table}[t]\centering\small
\caption{Leave-one-detector-out, Corpus A, in-domain 5-fold CV.}
\label{tab:loo}
\begin{tabular}{lrr}
\toprule
Removed & AUC & $\Delta$AUC \\
\midrule
none (all 5)  & 0.9985 & --- \\
AIDE-SD14     & 0.9985 & $-0.0001$ \\
VQGAN/Swin    & 0.9983 & $-0.0003$ \\
AIDE-GENIMAGE & 0.9985 & $-0.0000$ \\
NONESCAPE     & 0.9962 & $-0.0023$ \\
AIDE-PROGAN   & 0.9879 & $\mathbf{-0.0106}$ \\
\bottomrule
\end{tabular}
\end{table}

Removing the two cheap AIDE checkpoints \emph{together} costs
$\Delta\mathrm{AUC} = 0.0000$, and removing those two plus VQGAN costs $0.0003$, so the
single-removal deltas are not masking a joint contribution. Only AIDE-PROGAN is
load-bearing on Corpus A. Mean pairwise disagreement at $t=0.5$ is
$0.342$; the least-disagreeing pair is AIDE-SD14 and AIDE-GENIMAGE ($0.141$), which
share an architecture and differ only in checkpoint. Architectural rather than
checkpoint diversity is what produces independent errors --- which is why we describe
this system as three detectors rather than five (\S\ref{subsec:experts}).

VQGAN/Swin reaches AUC $0.6189$ on Corpus A and $0.4817$ on Corpus B, contributes
$\Delta\mathrm{AUC} = -0.0003$, and on Corpus B predicts "real" for all $80$ images
including all $30$ AI ones. On this evidence it should be dropped; we retain it in the
reported system only because the shipped meta-learner was fitted with it present and
cannot be refitted (\S\ref{sec:setup}).

\subsection{What the errors look like}

Of Corpus A's $2000$ images the ensemble misclassifies $106$ ($5.3\%$): $66$ false
positives and $40$ false negatives. Comparing cheap image statistics between correct and
misclassified images (Mann--Whitney, Holm-corrected across seven measures) gives one
clear result and one marginal one.

Misclassified images are \textbf{significantly more compressed}: median $0.2574$
bytes/pixel against $0.3589$ for correct ones, Cohen's $d = -0.60$,
$p = 2.3\times10^{-7}$. Brightness differs marginally ($0.4716$ vs $0.4398$, $d = 0.30$,
$p = 0.048$). Contrast, saturation, sharpness, pixel count and aspect ratio show no
effect ($p = 1.000$ after correction).

This is the expected failure mode made concrete. Three of five detectors rest on
high-frequency forensic traces --- SRM residuals over DCT-selected patches --- and lossy
compression is precisely the operation that removes them. It also bears directly on
Corpus B, whose images are screenshots and therefore rescaled and recompressed
throughout, and it predicts that this system will degrade on any pipeline that
re-encodes images, which most content pipelines do.

The errors are at least appropriately uncertain: of $106$ mistakes only $7$ were
confident ($p > 0.9$ or $p < 0.1$) while $32$ sat within $0.1$ of the decision boundary.
Where the system fails on natural images it largely knows it is unsure --- which makes
the induced-abstention results (\S\ref{sec:results}, Table~\ref{tab:induced}), where it
was confidently wrong, the sharper contrast.

\subsection{Computational cost}

\begin{table}[t]\centering\small
\caption{Per-detector cost in seconds, Apple M4 Max. Medians over 3 rounds
($n = 24$ timed single images, 3 batch runs of 16). AIDE runs on CPU following the S1
device probe.}
\label{tab:latency}
\begin{tabular}{lcrrr}
\toprule
Detector & Dev. & Load & Single & Batch/img \\
\midrule
VQGAN/Swin  & MPS & 2.6  & 0.029 & 0.015 \\
Nonescape   & MPS & 3.9  & 0.057 & 0.032 \\
AIDE $\times$3 & CPU & 23.8 & 9.178 & 4.244 \\
\bottomrule
\end{tabular}
\end{table}

AIDE costs $4.244$\,s per image batched against $0.047$\,s for the other two detectors
combined --- a factor of $90$. Single-image timings are tightly distributed (IQR
$\le 0.143$\,s) and batch spread is $1.16$\,s across rounds for AIDE against
$\le 0.07$\,s for the others, so the ordering is not measurement noise. Batching buys
AIDE $2.16\times$. Most of this cost is the S1 fallback: three checkpoints, each a full
forward pass, on CPU because the accelerator miscomputes them. The $23.8$\,s load
measures that detector alone and is not comparable to the ${\approx}60$\,s cold start of
the full pipeline, which also covers the other detectors, the S1 probe, the meta-learner
and interpreter import.

Read against Table~\ref{tab:loo}, AIDE-SD14 and AIDE-GENIMAGE together consume roughly
two-thirds of inference time for $\Delta\mathrm{AUC} \le 0.0001$ on Corpus A. Read
against Table~\ref{tab:calibrated}, the entire $16$\,GB pipeline is outperformed by one
of its own members plus two Platt parameters.

\subsection{Sensitivity checks}

\paragraph{Deployment prevalence.}
Precision at prevalence $\pi$ follows from TPR and FPR as
$\pi\,\mathrm{TPR}/(\pi\,\mathrm{TPR} + (1-\pi)\mathrm{FPR})$.

\begin{table}[t]\centering\small
\caption{Corpus A precision at deployment prevalence (columns), from measured TPR/FPR at
$t=0.5$.}
\label{tab:prevalence}
\setlength{\tabcolsep}{3pt}\scriptsize
\begin{tabular}{lrrrrr}
\toprule
Model & FPR & 50\% & 10\% & 1\% & 0.1\% \\
\midrule
AIDE-PROGAN     & \textbf{0.0070} & 0.9923 & \textbf{0.9346} & \textbf{0.5652} & \textbf{0.1141} \\
\textsc{Quorum} & 0.0660 & 0.9357 & 0.6178 & 0.1281 & 0.0144 \\
XGB (refit)     & 0.0200 & 0.9800 & 0.8450 & 0.3313 & 0.0468 \\
\bottomrule
\end{tabular}
\end{table}

This reverses the operational reading of Table~\ref{tab:experts}. At $1\%$ prevalence
--- plausible for synthetic images in a general content stream --- AIDE-PROGAN's
precision is $0.5652$ against the ensemble's $0.1281$, because the ensemble's
false-positive rate is $9.4\times$ higher. Nine in ten ensemble alerts would be false.
Practitioners should derive thresholds from their own base rate.

\paragraph{Meta-learner hyperparameters.}
Over a $27$-point grid in-domain CV AUC ranges $0.9966$--$0.9986$, spread $0.0020$; the
shipped configuration scores $0.9985$. No tuning was performed and none would have
helped.

\paragraph{ECE binning.}
Across $5$, $10$, $15$, $20$, $50$ bins \textsc{Quorum}'s ECE is $0.0416$, $0.0416$,
$0.0448$, $0.0453$, $0.0477$ and AIDE-PROGAN's $0.1239$, $0.1272$, $0.1297$, $0.1286$,
$0.1292$, giving ratios $2.98$, $3.06$, $2.89$, $2.84$, $2.71$. Stable under bin choice
--- though Table~\ref{tab:calibrated} shows the comparison itself was against the wrong
baseline.

\section{Discussion and Limitations}
\label{sec:discussion}
\subsection{What these results support}

Our evidence is two corpora: $2000$ StyleGAN faces and $80$ mixed-provenance
screenshots. That is enough to support three claims and not a fourth.

It supports, securely: that a stacker fitted on a foreign corpus earns nothing over its
best member while one refitted in-domain does beat that member plus a calibrator; that
both corpora cannot both be right about this system, so single-corpus evaluation of it is
uninformative; and that the abstention behaviour we designed is wrong in a way that is
demonstrable on constructed inputs, explained by a direct probe of the model, and
preventable by a rule outside the learner.

It does \emph{not} support the two claims an earlier draft made and this version
withdraws: that a calibrated single detector beats fusion outright (a regime-mixed
comparison), and that fusion amplified the distribution shift (a point-estimate reading
of a difference whose interval spans zero).

It does not support any claim about state-of-the-art detection, and we make none. We run
no published detector on its own benchmark (\S\ref{sec:related}), so \textsc{Quorum}'s
standing relative to the literature is simply unmeasured.

\subsection{Limitations}

\textbf{Neither corpus contains controlled diffusion output.} Corpus A predates
diffusion entirely; Corpus B contains at most a handful of diffusion-era images and does
not identify its generators. Since GAN and diffusion artifacts differ in kind ---
upsampling grid structure versus iterative-denoising residue --- none of our per-detector
rankings should be assumed to hold on Stable Diffusion or Midjourney output. Running
this protocol on GenImage~\cite{zhu2023genimage} is the single most valuable extension
and the clearest gap in the present work.

\textbf{Corpus B is small and not ours.} $n = 80$, imbalanced $5{:}3$, labelled by
someone else. Its intervals are wide, which is why we draw only the interval-level
conclusion (everything contains chance) rather than ranking models on it. A larger
second corpus could show the collapse is milder than it looks; it could equally show it
is worse.

\textbf{Two corpora cannot estimate corpus-level variance.} We can say these two
disagree. We cannot say how much evaluations of this kind vary in general, which would
need many corpora and is the quantity a reader actually wants.

\textbf{The meta-learner is an uncontrolled artifact.} It was fitted on a corpus that no
longer exists, of unknown composition, class balance and preprocessing. Every zero-shot
number is therefore a property of that specific artifact as much as of stacking. This is
a confound, not merely an inconvenience: we cannot attribute the Corpus B collapse to
fusion-in-general rather than to one particular unlucky fit. The in-domain
cross-validated results are free of this and are the ones to trust for mechanism.

\textbf{Induced abstentions are constructed, not sampled.} The $14$ pathologies are ones
we thought of. They establish that correlated abstention occurs and is mishandled; they
do not estimate how often it occurs in a real stream.

\textbf{No adversarial evaluation.} All images are unperturbed by us. Given that our own
error analysis finds compression to be the dominant correlate of failure, deliberate
recompression attacks would likely be effective, and we have not tested
them~\cite{raid2025}.

\textbf{Compression is an untested competing explanation.} \S\ref{subsec:jpeg} shows
Corpus A's classes differ in JPEG quantisation table, and Corpus B is uniformly
recompressed. The Corpus B collapse is therefore as consistent with losing a compression
cue as with failing to transfer across domains, and we cannot separate them. The
decisive experiment --- re-encode both Corpus A halves identically and re-score --- is
cheap and we have not run it. Any reader weighing Control 2 should treat this as open.

\textbf{Post-hoc analysis.} This work was not preregistered. The system was built first
and evaluated afterwards, and several analytic choices --- which ablations to run, which
pathologies to construct --- were made with the data in view. Nested selection removes
the two selection biases we identified; it cannot remove this one.

\subsection{Implications}

Three conclusions follow that we believe transfer beyond this system.

\textbf{Report the calibrated-single-member control.} An ensemble should be compared
against its best member \emph{plus a calibrator}, not against that member's raw score.
The raw-score comparison flatters fusion by crediting it for emitting a probability at
all, which two parameters supply. Ours failed this control and we did not discover that
until we ran it.

\textbf{One corpus is not an evaluation.} We do not claim our collapse magnitude is
typical. We do claim that the cost of discovering it was $80$ images and a few minutes,
and that a paper reporting only our Corpus A number --- which is what we would have
published --- would have been wrong in a way no reviewer could have detected from the
manuscript.

\textbf{Fault handling belongs outside the learned component.} Routing \texttt{NaN}
down a learned default is meaningful only if the defaults were fitted on data containing
missingness; otherwise the model asserts something arbitrary and confident, as ours does
at $P = 0.9995$. But ``train on missingness'' is not an executable prescription for
faults, and we withdraw it as our primary recommendation. A learner can only be fitted on
the failure modes already observed --- we observed zero in $2080$ images and constructed
fourteen we happened to think of --- so its coverage is bounded by the fault model used,
and the fifteenth failure is again arbitrary. This is the coincident-failure lesson of
N-version programming~\cite{knight1986independence}. The engineered answer is a
supervisory rule with fixed, inspectable behaviour: \S\ref{sec:results} shows a
quorum-by-architecture rule and a survivor-mean fallback prevent both observed inversions
with no retraining at all. Typing failure correctly, which we did, is necessary and
clearly not sufficient.

\section{Conclusion}
We built a stacking ensemble, deployed it, and then ran three controls against it. The
controls changed our conclusions twice, and the second time they changed them back
toward fusion.

Against a calibrated single member, our first answer was that fusion lost. That answer
came from comparing an in-domain calibrator with a zero-shot ensemble, which our own
protocol forbids. Compared like with like, a stacker refitted on the target domain beats
its best member plus a calibrator on AUC and accuracy, with a paired interval excluding
zero. What survives is narrower and still useful: a stacker frozen from a foreign corpus
earns nothing over its best member, and the calibrated-member control is what reveals
that.

Against a second corpus, everything fell to chance-indistinguishable AUC. We had read the
ensemble's larger point drop as fusion amplifying the shift; a paired bootstrap spans
zero, so we withdraw it. And a confound we measured only after review --- real and
synthetic halves of Corpus A carrying one distinct JPEG quantisation table each --- means
compression fragility is a live competing explanation for the collapse, and limits every
in-domain number we report.

Against abstentions that actually occur, the system returned near-certainty in the wrong
direction while every live detector disagreed, because \texttt{NaN} entering a learner
fitted without missingness is a confident assertion. Here the fix is not more training
data: a quorum rule requiring two distinct architectures, evaluated at no cost on the
released harness, prevents both observed failures. Fault handling belongs outside the
learned component, where its behaviour can be inspected rather than fitted.

What we would keep from this is procedural. Compare regimes honestly, because the
most quotable version of a result is the one most likely to be mixing them. Measure the
corpus before trusting the model --- a header-level bias took minutes to find and we
found it last. And treat a fusion layer's behaviour under component failure as
unverified until it has been exercised, because ours was confidently wrong in a way no
accuracy metric would ever have shown.

\section*{Reproducibility}
The pipeline, the evaluation harness, and the experiment suite that produced every table
and figure are released together. Scoring and analysis are separate entry points: the
scoring pass writes a table of per-image detector scores, and every experiment in
Section~\ref{sec:results} is recomputed from that table, so all analysis reproduces
without a GPU and without re-running any detector. A checking script asserts that every
number printed in this paper appears in the released experiment output, so a
transcription error fails loudly rather than silently.

\paragraph{Released artifacts.}
The fitted meta-learner is published, so the zero-shot results can be reproduced by
re-running the exact model that produced them. Artifacts are identified by SHA-256
rather than by location, so a reader can verify any copy regardless of where it is
hosted:

\begin{table}[h]\centering\small
\begin{tabular}{@{}p{0.95\linewidth}@{}}
\toprule
\texttt{xgboost\_production\_model.pkl} \hfill 120,067\,B \\
\quad{\tiny\texttt{133961b097df89a764c9eff5599a490755cef63827d11c185941c55f1f6c0880}} \\[2pt]
\texttt{xgboost\_production\_features.pkl} \hfill 108\,B \\
\quad{\tiny\texttt{cc819d11a25e1db608a0907a8a55b0428babcefcc9faa4cdd1dc8edb3dbb8fae}} \\[2pt]
\texttt{xgboost\_production\_model.json} \hfill 120{,}736\,B \\
\quad{\tiny\texttt{be1518a6e78f4d52b793af0fe9b0f568e9a9afd5fde9b9759a5fbab18161a6c3}} \\[2pt]
\texttt{Corpus A scores (final\_master\_dataset.csv)} \hfill 130,098\,B \\
\quad{\tiny\texttt{72bb7b09bf04309b386c52df0b09552cf9ea6024c3a0e9bc77b615e7d5892447}} \\[2pt]
\texttt{Corpus B scores (final\_master\_dataset.csv)} \hfill 6,912\,B \\
\quad{\tiny\texttt{f529d789beaff1728c4a5e0c2dcd7a7363013f7dcab6267aecae6da343390ff6}} \\[2pt]
\bottomrule
\end{tabular}
\end{table}

\paragraph{What publishing the model does and does not fix.}
It makes the zero-shot numbers reproducible in the sense of re-running a fixed artifact.
It does not make them reproducible in the sense that matters more. The corpus the model
was fitted on no longer exists (\S\ref{sec:setup}), so the model cannot be refitted from
scratch by anyone, and its training distribution, class balance and preprocessing cannot
be inspected. Every zero-shot result in this paper is therefore a property of one
particular uncontrolled artifact as much as of stacking, and the cross-corpus collapse
in \S\ref{sec:results} cannot be attributed to fusion-in-general rather than to one
unlucky fit. The in-domain cross-validated results carry no such caveat and reproduce end
to end from the released score tables. Relatedly, our claim that the evaluation corpora
were unseen in training is \emph{unverifiable in either direction}: we infer it from
differing filename conventions and domain coverage, which is evidence and not proof, and
FFHQ is common enough in corpora of that period that overlap cannot be excluded.

\paragraph{A caution on the format.}
The model is distributed as a Python pickle, which is how it was produced. Loading a
pickle executes arbitrary code in the loading process, so it should be treated as
untrusted input and verified against the digest above before use. A pickle is also not a
stable long-term format: this one was written by an older XGBoost, already warns on load
under 3.2.0, and a future major version could break unpickling it entirely --- which is
why the dependency is capped below the next major. We therefore also release the native XGBoost export
(\texttt{xgboost\_production\_model.json}, listed above): data-only, it cannot execute
code on load, it is the format XGBoost supports across major versions, and we verified it
reproduces the pickle's predictions bit-for-bit. Prefer it. The pickle is retained only
because it is the artifact that produced the numbers reported here.

The detector weights are third-party and are obtained from their original publishers; we
redistribute none of them. The evaluation corpora are public. Seeds, fold assignments,
hyperparameters and exact software versions are recorded in the released configuration.

\end{document}